# FINGERPRINT ORIENTATION REFINEMENT THROUGH ITERATIVE SMOOTHING

Pierluigi Maponi, Riccardo Piergallini and Filippo Santarelli

Department of Mathematics, University of Camerino, Camerino, Italy

## ABSTRACT

*We propose a new gradient-based method for the extraction of the orientation field associated to a fingerprint, and a regularisation procedure to improve the orientation field computed from noisy fingerprint images. The regularisation algorithm is based on three new integral operators, introduced and discussed in this paper. A pre-processing technique is also proposed to achieve better performances of the algorithm. The results of a numerical experiment are reported to give an evidence of the efficiency of the proposed algorithm.*

## KEYWORDS

*Fingerprint analysis, Equalisation, Segmentation, Orientation Extraction, Orientation Refinement*

## 1. INTRODUCTION

In the last few decades the interest in biometrics of government institutions, industries and scientific communities has grown very much. Many disciplines are engaged in the study and analysis of biometric traits: chemistry, engineering, law and mathematics are just few examples of the interested fields.

Among the other traits, fingerprints are one of the most widely used; this is due to their low storage requirements, their low cost acquisition systems and their distinctiveness; indeed, even two twins have different fingerprints. Comparing two fingerprints without the aid of modern technologies requires high skills and a large amount of time, even for trained people; thus leading to high costs. Recent advances in computer science and improvements of hardware performances allowed the development of several automated fingerprint recognition systems.

There are mainly two kinds of recognition systems: verification and identification. The former is mostly used in civilian applications, for instance in restricted resources access control, where the acquired fingerprint is compared to the ones already enrolled in the database and only two possible responses are expected: access granted or access denied. Identification is aimed to find the identity of a person, given its fingerprint; this is mainly used by law enforcement agencies for investigation purposes and the analysis of the crime scene.

Fingerprint images consist of an alternation of dark and bright curves, which are usually called ridges and valleys, respectively. Most recognition systems compare the position and orientation of the minutiae, that is either a termination of a ridge or a bifurcation of a ridge. The typical functioning scheme of a recognition system is:

1. Acquisition by digitalizing an inked paper or via direct scan of the finger.

2. Pre-processing, useful to enhance both contrast and brightness of input images, and to reduce the noise introduced by the acquisition phase.

3. Feature extraction, where the system computes some particular features, that will be used in the following stages.

4. Post-processing, a procedure that improves the extracted features by removing the worst features and codifying the right ones.

5. Matching, where a comparison between the newly acquired image and a database establishes the identity of the fingerprint owner, or grants access to restricted resources.

The orientation field is one of the most commonly extracted features, since it used for many purposes: classification [4], detection of singular points [5], detection of fingerprint alterations [6], registration before matching [7], matching performance improvement[8,9] and as a matching feature in itself[10]. So, the computation of the orientation field is a very important part in any recognition system and thus it is one of the deeply studied phases; many techniques have been developed both for the feature extraction and for the post-processing. Classical approaches to the orientation field computation are gradient-based[11], slit- and projection-based techniques[12] [13], frequency domain orientation estimation[14]. Regardless of the chosen approach, creases, scratches, discontinuous ridge patterns and no-signal areas brings very noisy areas in the acquired images, thus making necessary to post-process the extracted orientation field[3].Several interesting techniques to improve the orientation field reliability exist in literature, such as: orientation regularisation by using coherence criteria [15], neural network classification of unreliable orientations [16], multi-scale analysis for the correction of elements that change among different scales [17], enhancement by exploiting a global orientation model[18] and probabilistic approaches[19].

This paper describes an algorithm for pre-processing step, the orientation field extraction and proposes a novel approach for the field regularisation. In particular, we use a bank of directional gradient filters to compute the local orientation of the ridge flow: the image is convolved with each filter and the magnitude of its response gives a measure of the reliability of the corresponding orientation. We combine the information coming from every filter by a weighted average to achieve a reliable field of orientations. The regularisation procedure is mainly based on three operators that can be directly applied to the orientation field to produce a different kind of local improvements. The proposed regularisation algorithm combines these various local improvements to achieve a global enhancement of the orientation field.

A detailed description of our algorithm can be found in Section2; in Section3 we provide some experimental results, while in Section 4 we give some conclusions and future developments of this work.

## 2. ALGORITHM

We start with the description of the direction field and the orientation field of a fingerprint. We use the same definition proposed in Sherlock and Monro[21]; here we report this definition for the convenience of the reader.

Let $z \in \mathbb{C}$ be a complex number and $\theta \in [0,2\pi)$ be its argument. The direction associated to $z$ is given by the angle $\theta$; hence we can think the set of all the possible directions as the unit circle $S^1$. We can instead consider the straight line $tz$, with $t \in \mathbb{R}$, as a unique entity; so, it identifies a direction $\theta$ with its opposite $\theta+\pi$, and defines this entity as the orientation associated to $z$. Since $tz$ is invariant by rotations through integer multiples of $\pi$, the set of all the possible orientations can be identified with the projective circle $P^1$. When we need to describe the ridge flow of a fingerprint a direction field, namely a mapping from the image to $S^1$, is unsuitable due to the discontinuities that arise, for instance, along the symmetry axis of a loop (see Figure 1).

Conversely, using an orientation field, i.e. a mapping from the image to $P^1$, the two troublesome directions are identified and the field is continuous.

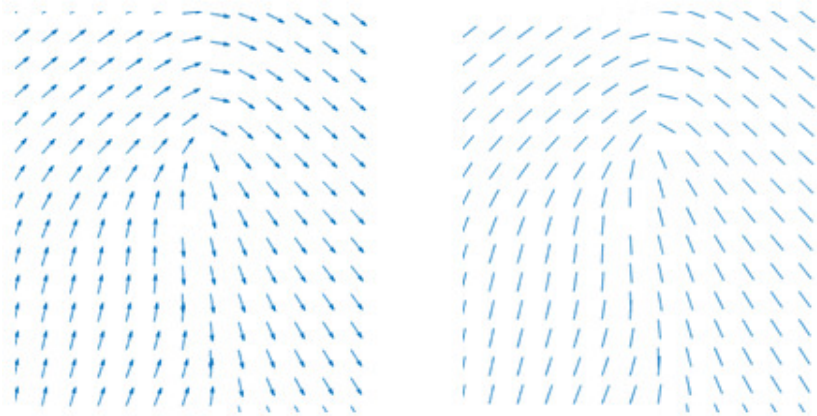

Figure 1 – Comparison between a vector field on the left and an orientation on the right.

We propose an algorithm to compute and refine the orientation field associated to the ridge flow of a given fingerprint. This algorithm is divided in the two stages: the pre-processing procedure, described in Section 2.1, that enhances the ridge-valley structure and even some image properties like contrast and brightness; the orientation extraction, outlined in Section0, that is the main step of our algorithm, when the orientation field is computed and refined.

## 2.1 Pre-processing

This is an important step of the algorithm, since it yields to homogeneous images with similar contrast and brightness, and produces the foreground mask, that allows the distinction of the fingerprint from background pixels. In the following, we refer to masks as matrices with values 0 or 255; when element-wise logical operations are performed between masks, we consider their 0 values as false entries, and their 255 values as true ones.

The pre-processing procedure consists of three sub-steps: equalisation, discussed in Section 2.1.1, segmentation, outlined in Section 2.1.2, and ridge amplification, described in Section 2.1.3.

### 2.1.1 Image Equalisation

Let I bean $N \times M$ image whose elements $I(i,j)$ with $i = 1, ..., N$ and $j = 1, ..., M$ range from 0 to $L - 1$ (when $I$ is an 8-bit image, $L = 256$); additionally suppose that local minima correspond to ridges and local maxima to valleys. We start with a linear scaling of $I$ in order to use the full grey level range. Then the histogram of the scaled image is convolved with a Gaussian kernel to improve its smoothness, and a simple adaptive procedure is used to compute a boundary value between dark pixel values, i.e. ridges, and bright pixel values, i.e. valleys. The equalised image$I_E$ is computed by a piece-wise linear transformation that maps the threshold value in the middle of the histogram, and removes small grey values variations around 0 and $L - 1$ which are often useless, see [22] for a detailed description of this equalization step.

### 2.1.2 Image Segmentation

The removal of the border is a fast preliminary step for the enhancement of the efficiency of the segmentation phase, since it reduces the area to process and it removes noisy parts often appearing on the border.

The very first operation consists of analysing rows and columns of image, starting from the image border to the image centre, computing their grey level variation and, if not sufficiently high, setting to false the corresponding line in the foreground mask $M_F$. More precisely, we set a

```
M_F :=matrix of 255s of the same size as I_E
for each row or column vector l
        V := max_j l(j) − min_j l(j)
        if V < τ_V then
                fill with 0s the corresponding row or column of M_F
        end if
end for
I_E' :=convolution of I_E with K_O
// Compute the coefficients of the top and bottom oblique line
{(m_t,q_t),(m_b,q_b)} :=line-fitting algorithm applied to I_E'
for each point (i,j) such that (i < m_t j + q_t) ∨ (i > m_b j + q_b)
        M_F(i,j) := 0
end for
// Repeat the operations for nearly vertical lines
I_E' :=convolution of I_E^T with K_O^T
{(m_t,q_t),(m_b,q_b)} :=line-fitting algorithm applied to I_E'
for each point (i,j) such that (i < m_t j + q_t) ∨ (i > m_b j + q_b)
        M_F^T(i,j) := 0
end for
```

Algorithm 1 – Pseudo-code for the border removal procedure.

minimum-allowed variation threshold $\tau_V$, then for each line scanned a maximum variation $V$ is computed and if $V < \tau_V$ the corresponding line in the foreground mask is set to false.

Since images often come from old fingerprint cards, they can present oblique lines in the image background, due to camera misalignment. These kind of lines, and the image part not relevant for the fingerprint analysis is removed from the foreground mask. Note that, $M_F$ accounts for the relevant part of the image and it is computed in the following procedure. We define a $5 \times \left[\frac{M}{4} + 1\right]$ kernel $K_O$, as follows

$$K_O = (1,1,0,-1,-1)^T \cdot (1,1,\ldots,1), \tag{1}$$

where the operator T stands for the matrix transposition. When $K_O$ is convolved with the image $I_E$, we get high responses at nearly horizontal lines; hence a line-fitting algorithm applied to the response matrix can find two lines: one near to the top border and another near to the bottom part of the image. So, we consider as background any pixel above the former line, and any pixel below the latter one. We can apply the same procedure using the kernel $K_O^T$ and the image $I_E^T$, to remove nearly vertical lines from of the foreground mask. A pseudo-code version of the border removal procedure is reported in Algorithm 1.

Handwritten text may appear in fingerprint cards; the next step aims to remove these signals, and other small artifacts. We apply a median filter with a circular structuring element to balance the width of ridges and valleys, and a Gaussian filter that removes sharp transitions of grey level values. A minimum filter is then applied to the resulting image to enlarge the ridges; moreover the image is scaled to fit the range $[0,1]$, squared and scaled again to the range $[0, L − 1]$, with the same purpose.

After this filtering step, a mask $M_0$ is created with true values where the image is under a fixed threshold $\tau_0$ and false values elsewhere. The mask $M_0$ is combined with the foreground mask $M_F$ computed in border removal procedure, by applying the element-wise logical conjunction ∧ between them. The resulting mask is dilated to fill up the valleys, it is eroded to remove spurious parts, and the majority of its connected components are filtered out, keeping only the largest ones, where the holes are filled up. A final dilation is then performed to compensate for the previous strong erosion; the resulting mask is denoted with $M_1$.

The reliability of the foreground mask is improved by using the information coming from the alternation of ridges and valleys; in particular, an edge detector, such as the well-known Canny edge detector, can highlight sharp transitions of the grey levels, producing a lot of lines close to each other where ridges and valleys alternate; elsewhere only small and isolated lines are created. Let $M_2$ be the mask produced by the edge detector, and $I_2$ be the image obtained convolving $M_2$ with a Gaussian kernel; we keep only the connected components of $M_2$ with sufficiently high mean intensity values, where the values are taken from $I_2$. The resulting mask is dilated, eroded, filtered and dilated again as we did for $M_1$; let $M_2$ be the final mask.

The information coming from the two masks $M_1$ and $M_2$ is brought together by logical disjunction; the convex hull of the resulting mask is computed giving the desired foreground mask, that we keep calling $M_F$.

### 2.1.3 Ridges amplification

Ridges and valleys are not always clearly separated; in the following we define a procedure to locally enhance the difference between them. Let $I_R$ be the restriction of the equalised image $I_E$ to the significant mask $M_F$ obtained in the segmentation step. We compute $I_M$ by applying a circular maximum filter to $I_R$; in the same way we compute the image $I_m$, applying a circular minimum filter to $I_R$. Let $t_{M_1}, t_{M_2} \in [0, L-1]$ be two threshold values, a linear transformation mapping $t_{M_1}$ to 0 and $t_{M_2}$ to 255, is applied to $I_M$ to obtain a scaled image $I_{M,2}$. In a similar way, from the thresholds $t_{m_1}$ and $t_{m_2}$ is obtained $I_{m,2}$ from $I_m$. So each value of the initial image can be normalized to the new local minimum and maximum, and the final image $I_A$ can be computed as follows:

$$I_A(i,j) = I_{m,2}(i,j) + \frac{I_R(i,j) - I_m(i,j)}{I_M(i,j) - I_m(i,j)}(I_{M,2}(i,j) - I_{m,2}(i,j)). \qquad (2)$$

A demonstration of the efficacy of our pre-processing algorithm can be seen in Figure 2 and Figure 3, where the initial input image is shown on the left, and the final output on the right.

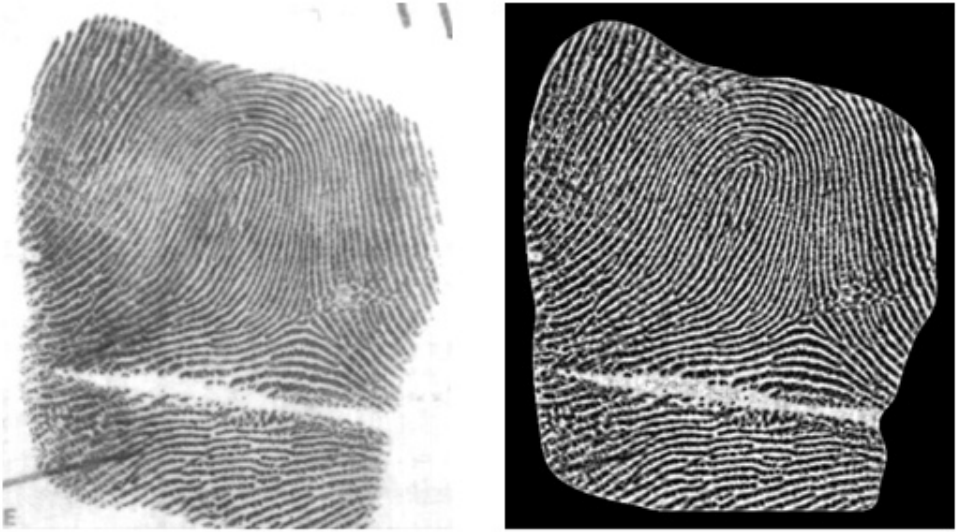

Figure 2 – An application of our pre-processing technique to a fingerprint image, on the left; the right image is the algorithm output.

## 2.2 Orientation Extraction

For the sake of simplicity, we denote $I$ the image computed, in the pre-processing stage, that is the image denoted $I_A$ in Section 2.1.

The orientation extraction procedure is composed of three steps: the orientation estimation, described in Section 2.2.1; the spatial period computation outlined in Section 2.2.2; and finally the orientation refinement, described in Section 2.2.3.

### 2.2.1 Orientation Estimation

The estimation of the local orientation field is performed using a bank of directional gradient filters; each filter is composed by two components: a Gaussian in a given direction, and a Gaussian derivative in the orthogonal direction. More precisely, given $r \in \mathbb{R}_+$, we define the function $K_{r,0}: [-r, r]^2 \to \mathbb{R}$:

$$K_{r,0}(s,t) = d(s) \cdot e^{-\left|\frac{d(s)}{\sigma_1}\right|^{2\alpha_1}} \cdot e^{-\left|\frac{d(t)}{\sigma_2}\right|^{2\alpha_2}}, \qquad (3)$$

where $d(t) = t - \left(\frac{r}{2} + 1\right)$, and $\sigma_1, \alpha_1, \sigma_2, \alpha_2 > 0$ are shape parameters. We select $N_A$ equally spaced angles $\theta_1, \ldots, \theta_{N_A} \in [0, \pi)$ and define the following group of directional gradient kernels:

$$K_{r,k}(i,j) = K_{r,0}(i\cos\theta_k + j\sin\theta_k, -i\sin\theta_k + j\cos\theta_k), \qquad (4)$$

Where $k = 1, \ldots, N_A$, $i, j = 1, \ldots, r$. The response to the $k$-th filter gives the directional image derivative along the direction with angle $\theta_k$.

An orientation estimation is extracted as follows: we convolve the image $I$ with each kernel $K_{r,k}$, compute the absolute value of the response, and smooth it with a Gaussian filter; we call $W_k$ the resulting matrix. Let $\phi_k = \left(\theta_k + \frac{\pi}{2}\right) \mod \pi$; $\mathcal{O}_k = W_k e^{i\phi_k}$ is a complex matrix with high-magnitude elements where ridges have orientation given by angle $\theta_k$. The desired orientation field $\mathcal{O}$ is computed as:

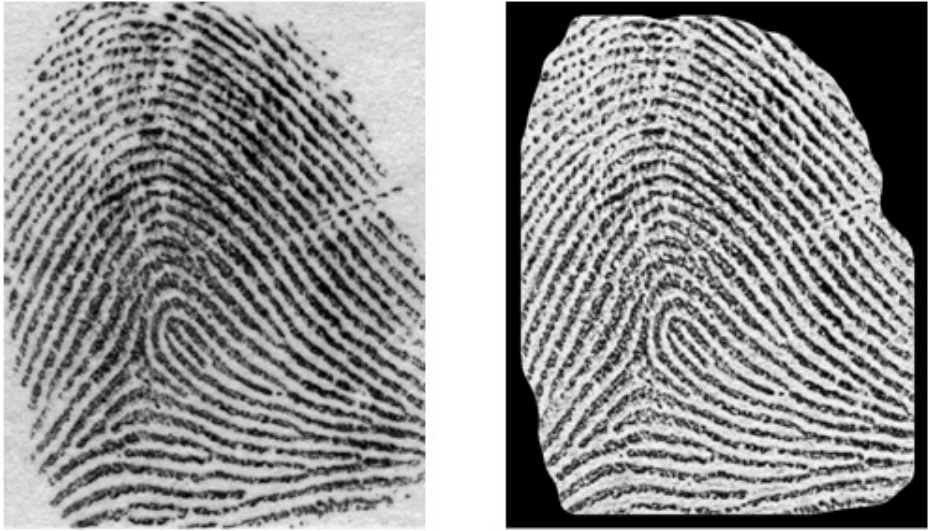

Figure 3 – Another example to assess the validity of the pre-processing technique outlined in Section 2.1.

$$\mathcal{O}(i,j) = \frac{\sum_{k=1}^{N_A} W_k(i,j) \cdot e^{i\phi_k}}{\sum_{k=1}^{N_A} W_k(i,j)} = \frac{\sum_{k=1}^{N_A} \mathcal{O}_k(i,j)}{\sum_{k=1}^{N_A} W_k(i,j)}. \tag{5}$$

To describe the local orientation field we have used the upper complex half-plane representation; indeed $\mathcal{O}$ maps each pixel to an element of $P_1$, identifiable with a complex number with phase angle in the interval $[0, \pi)$. Additionally, the magnitude of the complex number gives the reliability of the orientation in that point. As well pointed out by Kass and Witkin[23], when we perform operations among orientations, we have to convert them to a continuous vector field; in our case it is sufficient to square the orientations as complex numbers, thus doubling the angles, perform the necessary operations, and then take the square root of the result, that halves the angles and brings them back in the range $[0, \pi)$.

A Gaussian smoothing must be performed to remove sharp transitions, arising from the use of $N_A$ pre-defined orientations; the smoothing must be preceded by squaring and followed by the square root computation, as described before. Finally, an absolute value normalisation must be applied to the orientations field; this operation deals only with magnitudes and does not need to be surrounded by squaring operations.

### 2.2.2 Spatial Period Computation

From the orientation field $\mathcal{O}$ we can estimate the distance between two consecutive ridges. Let

$$\{(i_n, j_n) | n = 1, \ldots, N_P\} \tag{6}$$

be the set of the indices of entries in $\mathcal{O}$; note that, these entries are selected by choosing a uniform grid of $\mathcal{O}$. For each point $(i_n, j_n)$, we consider the fixed-length segment centred at $(i_n, j_n)$ and orthogonal to the orientation $\mathcal{O}(i_n, j_n)$; on this segment we select $N_S$ uniformly distributed points, and for each of these points an orientation $o_{n,k} \in \mathbb{C}$ and a value $v_{n,k} \in [0, L-1], k = 1, \ldots, N_S$, are obtained by respectively interpolating the field $\mathcal{O}$ and the image $I$.

In order to consider the $n$-th segment sufficiently reliable, a minimum threshold $\tilde{t}$ is chosen and the condition $\min\{|o_{n,k}|\}_{k=1,\ldots,N_S} > \tilde{t}$ must be fulfilled, otherwise the segment is skipped. Let us suppose that the $n$th segment is reliable, we consider the discrete signal $v_{n,k}, k = 1, \ldots, N_S$, to which a low-pass filter is applied. From the discrete Fourier transform of the resulting signal is computed the spatial frequency $f_{s,n}$ that corresponds to the first peak after the zero-frequency. The spatial frequency for the whole image can be computed as

$$f_s = \frac{1}{N_P} \sum_{n=1}^{N_P} f_{s,n}. \tag{7}$$

The spatial period $T_s$, i.e. the distance between two consecutive ridges, is given by $T_s = \frac{N_S}{f_s}$.

### 2.2.3 Orientation Refinement

Let $F$ be a rectangular region in the image $I$, and $\mathcal{F}: F \to \mathbb{C}$ be an orientation field defined on $F$, possibly given by the initial estimation $\mathcal{O}$; we denote with $\mathcal{F}(\mathbf{x})$ the orientation $\mathcal{F}(x, y)$ at point $\mathbf{x} = (x, y)^T \in F$.

Our algorithm to refine the orientation field relies on three operators: adjuster, smoother, drifter. Let $\mathcal{C}_R = \{\mathbf{r}_k\}_{k=1,\ldots,N_C} \subset \mathbb{R}^2$ be a set of $N_C$ points selected from the circumference of radius $R \in \mathbb{N}$ centred at $\mathbf{0}$; for every $k = 1, \ldots, N_C$, let $p_k = \mathbf{r}_k \cdot \mathbf{i} + i\mathbf{r}_k \cdot \mathbf{j}$, where ı and ȷ are the usual vectors of the canonical base for $\mathbb{R}^2$, $i$ is the imaginary unit and · denotes the inner product. Let $\mathcal{G}_A$ be the orientation field:

Signal & Image Processing : An International Journal (SIPIJ) Vol.8, No.5, October 2017$$\mathcal{G}_A(\mathbf{x}) = \sum_{k=1}^{N_C} \operatorname{sgn}[w_k(\mathbf{x})] w_k{}^2(\mathbf{x}) \mathcal{F}(\mathbf{x} + \mathbf{r}_k), \tag{8}$$

where

$$w_k(\mathbf{x}) = \frac{\Re\left[\frac{\mathcal{F}(\mathbf{x}+\mathbf{r}_k)}{|\mathcal{F}(\mathbf{x}+\mathbf{r}_k)|} \frac{\overline{p_k}}{|p_k|}\right]}{\sum_{j=1}^{N_C} \Re\left[\frac{\mathcal{F}(\mathbf{x}+\mathbf{r}_j)}{|\mathcal{F}(\mathbf{x}+\mathbf{r}_j)|} \frac{\overline{p_j}}{|p_j|}\right]}, \tag{9}$$

$\overline{p_k}$ is complex conjugate of $p_k$, and $|\cdot|$ is the absolute value in $\mathbb{C}$. Essentially $\mathcal{G}_A(\mathbf{x})$ is the weighted sum of the field value over the circle of radius $R$ centred in $\mathbf{x}$; the weights are given by the function $w_k(\mathbf{x})$, and the sign function gives to $\mathcal{F}(\mathbf{x} + \mathbf{r}_k)$ the same orientation of the radial vector $\mathbf{r}_k$.

The adjuster $\mathcal{A}$ of the orientation field $\mathcal{F}$ is defined as:

$$\mathcal{A}_0[\mathcal{F}]^2(\mathbf{x}) = (1-s)\mathcal{F}(\mathbf{x})^2 + s\mathcal{G}_A(\mathbf{x})^2, \tag{10}$$

$$\mathcal{A}[\mathcal{F}](\mathbf{x}) = \frac{\mathcal{A}_0[\mathcal{F}](\mathbf{x})}{|\mathcal{A}_0[\mathcal{F}](\mathbf{x})|} \max(|\mathcal{F}(\mathbf{x})|, |\mathcal{G}_A(\mathbf{x})|), \tag{11}$$

where $s \in (0,1)$ is a relaxation parameter. In formula (10) the fields are squared before the addition; as above mentioned, this is a preliminary operation required to process any orientation field. The formula (11) gives a normalization of the adjusted field $\mathcal{A}[\mathcal{F}]$ in such a way that its magnitude is not less than the magnitude of the field $\mathcal{F}$.

The second operator, i.e. the smoother $\mathcal{S}$, is defined through operator

$$\mathcal{G}_S(\mathbf{x}) = \sum_{k=1}^{N_C} \operatorname{sgn}[u_k(\mathbf{x})] w_k{}^2(\mathbf{x}) \mathcal{F}(\mathbf{x} + \mathbf{r}_k), \tag{12}$$

where $w_k(\mathbf{x})$ is defined as in (9),

$$u_k(\mathbf{x}) = \frac{\Re\left[\frac{\mathcal{F}(\mathbf{x}+\mathbf{r}_k)}{|\mathcal{F}(\mathbf{x}+\mathbf{r}_k)|} \frac{\overline{\mathcal{F}(\mathbf{x})}}{|\mathcal{F}(\mathbf{x})|}\right]}{\sum_{j=1}^{N_C} \Re\left[\frac{\mathcal{F}(\mathbf{x}+\mathbf{r}_j)}{|\mathcal{F}(\mathbf{x}+\mathbf{r}_j)|} \frac{\overline{\mathcal{F}(\mathbf{x})}}{|\mathcal{F}(\mathbf{x})|}\right]}. \tag{13}$$

The sum in formula (12) is different from the one in (8), because the sign function now gives to $\mathcal{F}(\mathbf{x} + \mathbf{r}_k)$ the same orientation of $\mathcal{F}(\mathbf{x})$. The smoother $\mathcal{S}$ of the orientation field $\mathcal{F}$ is defined as:

$$\mathcal{S}_0[\mathcal{F}]^2(\mathbf{x}) = (1-s)\mathcal{F}(\mathbf{x})^2 + s\mathcal{G}_S(\mathbf{x})^2, \tag{14}$$

$$\mathcal{S}[\mathcal{F}](\mathbf{x}) = \frac{\mathcal{S}_0[\mathcal{F}](\mathbf{x})}{|\mathcal{S}_0[\mathcal{F}](\mathbf{x})|} \max(|\mathcal{F}(\mathbf{x})|, |\mathcal{G}_S(\mathbf{x})|). \tag{15}$$

For these relations holds an observation similar to the one made for relations (10) and (11).

The third operator, i.e. the drifter, is given in two versions: the tangent-weighted drifter $\mathcal{D}_T$ and the normal-weighted drifter $\mathcal{D}_N$; they are defined as follows:

$$\mathcal{D}_T[\mathcal{F}](\mathbf{x}) = \sum_{k=1}^{N_C} w_k(\mathbf{x}) \mathcal{F}(\mathbf{x} + \mathbf{r}_k), \tag{16}$$

36

Signal & Image Processing : An International Journal (SIPIJ) Vol.8, No.5, October 2017

$$\mathcal{D}_N[\mathcal{F}](\mathbf{x}) = \sum_{k=1}^{N_C} v_k(\mathbf{x})\mathcal{F}(\mathbf{x} + \mathbf{r}_k), \quad (17)$$

where $v_k(\mathbf{x})$ is defined as

$$v_k(\mathbf{x}) = \frac{\mathfrak{I}\left[\frac{\mathcal{F}(\mathbf{x}+\mathbf{r}_k)}{|\mathcal{F}(\mathbf{x}+\mathbf{r}_k)|} \frac{\overline{p_k}}{|p_k|}\right]}{\sum_{j=1}^{N_C} \mathfrak{I}\left[\frac{\mathcal{F}(\mathbf{x}+\mathbf{r}_j)}{|\mathcal{F}(\mathbf{x}+\mathbf{r}_j)|} \frac{\overline{p_j}}{|p_j|}\right]}, \quad (18)$$

and $\mathfrak{I}$ stands for the imaginary part of a complex number. The main difference between these operators is that $\mathcal{D}_N$ gives higher weights to field values normal to the radial vector, while $\mathcal{D}_T$ produces higher weights to the ones tangent to the radial vector.

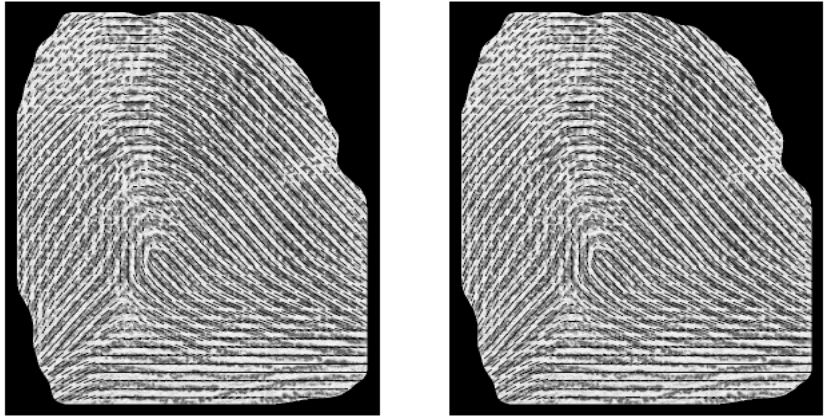

Figure 5 – The image on the left exhibits the orientation field computed by our algorithm, while the image on the right shows the output of its refinement

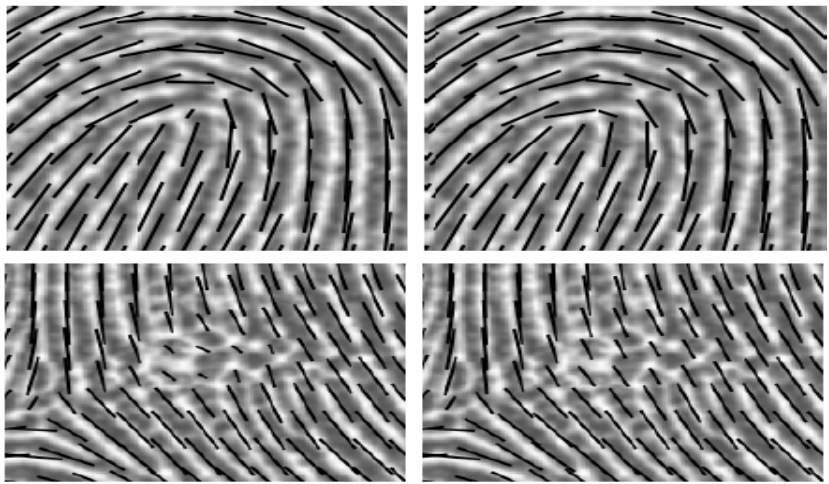

Figure 6 – Two magnified details from Figure 4; the extracted orientation is shown on the left column, its refined version in the right column

The smoother is actually the operator that enhances the orientation field, and gives global coherence even in the noisy parts. Applying it to the whole field may be dangerous, since it shifts loops ahead along their symmetry axis; hence we need a reliable mask that hides loops, to safely apply the smoother. In the following we will make use of the properties of all the three operators to create such a mask.

The extraction of the orientation field through directional gradient filters makes the loops shift ahead along the symmetry axis; the application of the smoother, as we mentioned before, emphasises that effect. The adjuster is very important because it has a rounding effect nearby the loops, and it is able to bring the loop centre back in the right position.

Suppose that $I$ is the image, $\mathcal{O}$ the orientation field, $M_F$ the foreground mask and $T_S$ the distance between two consecutive ridges obtained from the previous steps. The very first operation is a new estimation of the orientation field using directional gradient filters of radius $\frac{3}{2}T_S$; let $\mathcal{O}_1$ be the resulting field. Let $\rho_S > 0$, we apply the smoother, with radius $\rho_S T_S$, to globally improve the field, even with the already mentioned drawbacks. Let $\rho_{D_1} > 0$ be another scaling factor, $\mathcal{D}_T[\mathcal{S}[\mathcal{O}_1]]$ and $\mathcal{D}_N[\mathcal{S}[\mathcal{O}_1]]$ are computed, both with radius $\rho_{D_1} T_S$; both the tangent-weighted and the normal-weighted drifter have high magnitude on loops, highlighting different part of it, so if they are used together the loop position and extension are more accurate. A remarkable property in an orientation field is the duality between loops and deltas through complex conjugation; we will make use of it in the following. Let $\tau_1 > 0$ be a threshold value and $M_1$ be the mask defined as

$$M_1 = \frac{|\mathcal{D}_T[\mathcal{S}[\mathcal{O}_1]]| + |\mathcal{D}_N[\mathcal{S}[\mathcal{O}_1]]| - |\mathcal{D}_T[\mathcal{S}[\overline{\mathcal{O}_1}]]| - |\mathcal{D}_N[\mathcal{S}[\overline{\mathcal{O}_1}]]|}{2} > -\tau_1, \qquad (19)$$

where $\overline{\mathcal{O}_1}$ stands for the complex conjugate of the orientation field $\mathcal{O}_1$. The mask $M_1$ has false entries only on deltas. Finally a logical conjunction is applied to the mask and the initial mask $M_F$, to join the information coming from the segmentation step; morphological erosion is performed to be sure that deltas are fully uncovered. An image $S_1$ is computed from the mask through Gaussian blurring.

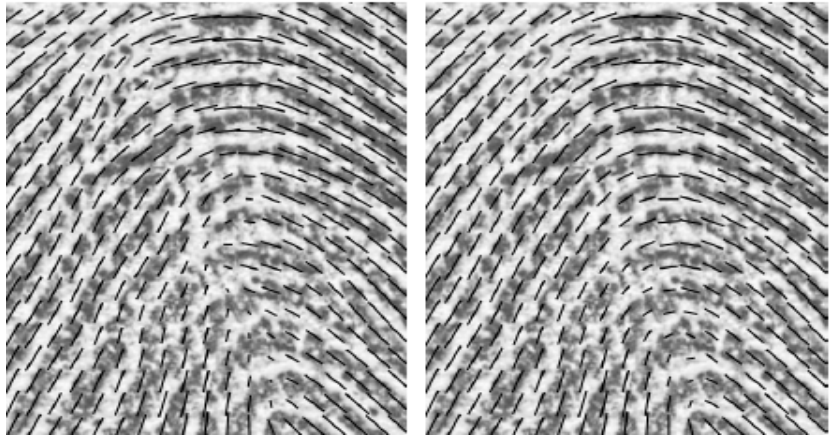

Figure 7 – Magnified version of two details taken from Figure 5

Another orientation field $\mathcal{O}_2$ is computed using a radius of $\frac{1}{2}T_S$; since $\mathcal{O}_2$ is less sensitive to orientation changes, the loop structure may be altered and we use the adjuster to recover it. Let $\mathcal{A}$ be the adjuster operator of radius $\rho_A T_S$. In this case, for each entry, we use a different strength value that is given by the strength image $S_1$; in this way the adjuster does not operate on deltas, preserving their integrity, while loops are improved. Let $\tau_2 > 0$ be a fixed threshold value and $M_2$ the mask defined as

$$M_2 = \frac{|\mathcal{A}[\mathcal{S}[O_1]] - \mathcal{A}[O_2]|}{2} > \tau_2. \tag{20}$$

The mask $M_2$ is then combined with the initial mask $M_F$ by logical conjunction, the smallest connected component are removed, and a final dilation is performed. This mask highlights the differences between the two orientation fields $\mathcal{S}[O_1]$ and $O_2$; these parts are very sensitive to changes in the radius used to compute the orientation fields, while an ideal fingerprint would not do that, thus these parts need to be improved by the smoother.

Now we get a more accurate mask with highlighted loops and deltas; we use the same idea as in (19), but using the adjusted field $O_3 = \mathcal{A}[\mathcal{S}[O_1]]$. So let

$$I_3 = \frac{|\mathcal{D}_T[O_3]| + |\mathcal{D}_N[O_3]| - |\mathcal{D}_T[\overline{O_3}]| - |\mathcal{D}_N[\overline{O_3}]|}{2}. \tag{21}$$

After having blurred the image with a Gaussian kernel, we can get a loop-delta mask as follows; let $\tau_3 > 0$ be a fixed threshold value, $M_0$ the erosion of the initial mask $M_F$, we compute

```
do
                         O' := S[O₃]
                         D = |O' - O₃|/2 > τ₄
    M₄ :=dilation of [(erosion of M₄)∧D]
    W :=Gaussian blurring of M₄ (assuming 0 on false entries, 1 on
         true)
       for each element at position (i,j)
                 O'(i,j) := O'(i,j)W(i,j) + O₃(i,j)(1 - W(i,j))
       end for
while M₄ is not empty
```

Algorithm 2 – Pseudo-code for the iterative smoothing

$$M_3 = (|I_3| > \tau_3) \wedge M_0. \tag{22}$$

Finally, the smallest connected components are removed from the mask $M_3$. The mask $M_3$ is used to remove singularities from the differences mask $M_2$ as follows

$$M_4 = M_2 \wedge \neg M_3, \tag{23}$$

where $\neg$ is the usual logical negation operator. A final dilation is performed on $M_4$ before starting the iterative smoothing procedure, so that the mask contains only the parts of the orientation fields that need to be improved and covers completely the singularities.

The last stage of the algorithm is an iterative smoothing procedure, where the smoother improves the orientation field over the mask until a halting condition is verified. In Algorithm 2 a pseudo code outlines our iterative smoothing algorithm.

Note that, at each step of this procedure the smoothing is performed, a difference mask with the previous field is computed and combined with the erosion of the previous mask through a logical conjunction to guarantee convergence; the algorithm stops when the current mask is empty.

A couple of examples are provided through Figure 4 and Figure 5, to witness the efficacy of our algorithm for the orientation refinement. In each figure, on the left the initial orientation field estimation is shown, while the results of the refinement procedure are exhibited on the right.

## 3. EXPERIMENTAL RESULTS

In this section some of the results obtained through our algorithm are shown; in particular we want to assess the validity of the orientation extraction and refinement procedure. Several images have been taken as input for our method and processed using the following parameters: $r = 15$, $\sigma_1 = 1$, $\alpha_1 = 2$, $\sigma_2 = 0.85$, $\alpha_2 = 2$, $N_A = 36$, $N_S = 31$, $\tilde{t} = 0.25$, $\rho_S = 1$, $\rho_{D_1} = 1$, $\tau_1 = 0.3$, $\rho_A = 0.7$, $\tau_2 = 0.5$, $\tau_3 = 0.1$. In this paper we show only four, that best illustrate the good behavior of our method.

Figure 6 shows two magnified details taken from Figure 4. In the upper row we can see the area surrounding the loop; before the refinement the orientation field does not follows the ridges where their curvature is very high, due to the use of directional gradient filters. However, the refinement procedure is able to reconstruct the real orientation field, shifting the loop centre back along its symmetry axis. Furthermore, we can see in the lower row of Figure 6 that the refinement algorithm is also able to recover the field in a noisy area, where the signal is very disturbed.

Another example of the benefits of our method comes from Figure 7, where a detail taken from Figure 5 is magnified. The left image presents the initial orientation field, estimated as outlined in Section 2.2.1; on the right image the corresponding refined orientation field is shown. Due to a crease and the use of directional gradient filters, the loop centre is far from its real position; the refinement procedure improves the field yielding very good results, even in this case.

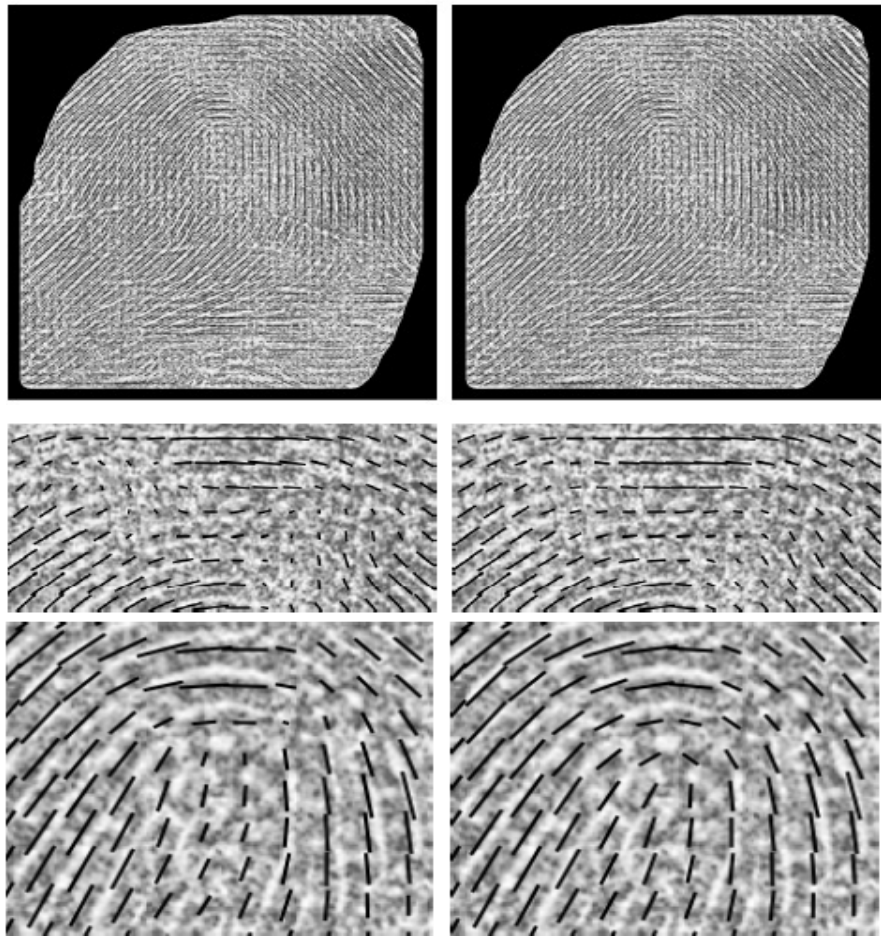

Figure 8 – Another example to assess the benefits of our refinement procedure. On the left column there are the orientation fields before the improvement, on the right column there are their refined counterparts

In Figure 8 we exhibit a fingerprint image that, despite the application of our refinement procedure, still presents some issues. In the first column of Figure 8 there are the orientation fields before the application of the regularisation algorithm, while in the second column there are their corresponding refined version. The first two images from the top show the fingerprint globally; if we focus on some of its details we can observe that around loops the refinement performs very good, while in very noisy parts it cannot fully reconstruct the real orientation field. Despite the behaviour is not optimal, the procedure improves the local orientation field even in those areas.

## 4. CONCLUSION

This paper describes a gradient-based procedure to extract the orientation field, and introduces a novel approach to the regularisation of the extracted one. Some results have been shown in Section 3 to support the efficiency and reliability of our method.

These results give an experimental evidence of the efficacy of the proposed algorithms; nevertheless, future investigations should be provided in order to assess precisely the algorithm performances. In particular, we will have to test our algorithm against a database with ground truth information, such as [24] and [25], and to compare this algorithm with other existing methods in terms of its accuracy and computational time.

Another interesting study regards the theoretical properties of the operators defined in Section 2.2.3 with respect to singular points and perturbations (such as image noise, misaligned orientation field and so on) in the fingerprint structure. It is also worth noting the use of the proposed algorithm in the authentication and identification applications also included the computation of fingerprint minutiae.

Signal & Image Processing : An International Journal (SIPIJ) Vol.8, No.5, October 2017[9] Tico, M., &Kuosmanen, P. (2003). Fingerprint matching using an orientation-based minutia descriptor. IEEE Transactions on Pattern Analysis and Machine Intelligence, 25(8), 1009-1014.

[10] Kulkarni, J. V., Patil, B. D., &Holambe, R. S. (2006). Orientation feature for fingerprint matching. Pattern Recognition, 39(8), 1551-1554.

[11] Bazen, A. M., &Gerez, S. H. (2002). Systematic methods for the computation of the directional fields and singular points of fingerprints. IEEE transactions on pattern analysis and machine intelligence, 24(7), 905-919.

[12] He, Y., Tian, J., Luo, X., & Zhang, T. (2003). Image enhancement and minutiae matching in fingerprint verification. Pattern recognition letters, 24(9), 1349-1360.

[13] Ji, L., & Yi, Z. (2008). Fingerprint orientation field estimation using ridge projection. Pattern Recognition, 41(5), 1491-1503.

[14] Chikkerur, S., Cartwright, A. N., &Govindaraju, V. (2007). Fingerprint enhancement using STFT analysis. Pattern recognition, 40(1), 198-211.

[15] Jiang, X., Liu, M., &Kot, A. C. (2004, August). Reference point detection for fingerprint recognition. In Pattern Recognition, 2004. ICPR 2004. Proceedings of the 17th International Conference on(Vol. 1, pp. 540-543). IEEE.

[16] Zhu, E., Yin, J., Hu, C., & Zhang, G. (2006). A systematic method for fingerprint ridge orientation estimation and image segmentation. Pattern Recognition, 39(8), 1452-1472.

[17] Oliveira, M. A., &Leite, N. J. (2008). A multiscale directional operator and morphological tools for reconnecting broken ridges in fingerprint images. Pattern Recognition, 41(1), 367-377.

[18] Chen, X., Tian, J., Zhang, Y., & Yang, X. (2006, January). Enhancement of low quality fingerprints based on anisotropic filtering. In International Conference on Biometrics (pp. 302-308). Springer, Berlin, Heidelberg.

[19] Lee, K. C., &Prabhakar, S. (2008, September). Probabilistic orientation field estimation for fingerprint enhancement and verification. In Biometrics Symposium, 2008. BSYM'08 (pp. 41-46). IEEE.

[20] Turroni, F., Maltoni, D., Cappelli, R., & Maio, D. (2011). Improving fingerprint orientation extraction. IEEE Transactions on Information Forensics and Security, 6(3), 1002-1013.

[21] Sherlock, B. G., &Monro, D. M. (1993). A model for interpreting fingerprint topology. Pattern recognition, 26(7), 1047-1055.

[22] Maponi, P.,Piergallini R.&Santarelli, F. (2017). Extraction and refinement of fingerprint orientation field. ICCSEA, WiMoA, SPPR, GridCom, CSIA, 57– 68.

[23] Kass, M., &Witkin, A. (1987). Analyzing oriented patterns. Computer vision, graphics, and image processing, 37(3), 362-385.

[24] Dorizzi, B., Cappelli, R., Ferrara, M., Maio, D., Maltoni, D., Houmani, N., ... &Mayoue, A. (2009). Fingerprint and on-line signature verification competitions at ICB 2009. Advances in Biometrics, 725-732.

[25] Watson, C. I. (2016). NIST Special Database 4. NIST 8-Bit Gray Scale Images of Fingerprint Image Groups. World Wide Web-Internet and Web Information Systems.
42

## AUTHORS

**Pierluigi Maponi** is associate professor in Numerical Analysis at the School of Science and Technology of Camerino University. Besides fingerprint analysis, his interest for image processing also concerns automatic age estimation, biomedical imaging and diagnostics, satellite imaging. Other research fields are numerical linear algebra, inverse problems and applications, computational fluid dynamics, hazard evaluation for water-related events.

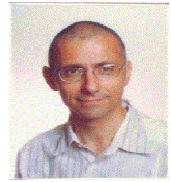

**Riccardo Piergallini** is full professor in Geometry at the School of Science and Technology of Camerino University. The main field of his scientific activity is low-dimensional topology. In particular, he is interested in the theory of branched coverings, as a tool for representing manifolds and studying various topological and geometric structures on them. Recently, he started to consider computational applications of topology and geometry, specially the ones concerning spatial modeling, image processing, computer graphics and artificial vision.

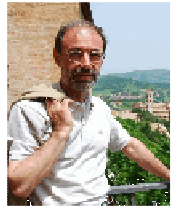

**Filippo Santarelli** is a PhD student at the School of Science and Technology of Camerino University. His research interests are fingerprint analysis and voice recognition.

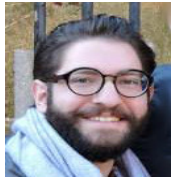